%% file: main.tex
\documentclass[letterpaper, 10 pt, conference]{ieeeconf}   
\IEEEoverridecommandlockouts 
\usepackage[utf8]{inputenc}
\usepackage{amsmath}
\usepackage{amssymb}
\usepackage{amsthm}
\usepackage{hyperref}
\usepackage[position=top]{subfig}
\usepackage{amsfonts}
\usepackage{mathtools}
\usepackage{url}
\usepackage{bm}
\usepackage{enumerate}
\usepackage{xcolor}
\usepackage{graphicx}
\usepackage{subfig}
\usepackage[font=small]{caption}

\title{2022-deformable-elastic-objects-UAVs}
\author{chiara gabellieri }
\date{May 2022}
\include{symbol}
\title{\LARGE \bf On the Existence of Static Equilibria of a Cable-Suspended Load with\\ Non-stopping Flying Carriers}   
\author{Chiara Gabellieri$^{*}$ and Antonio Franchi$^{*,\dagger}$
\thanks{$^*$ Robotics and Mechatronics Department, Electrical Engineering,  Mathematics, and Computer Science (EEMCS) Faculty, University of Twente, 7500 AE Enschede, The Netherlands. {\tt\footnotesize c.gabellieri@utwente.nl}, {\tt\footnotesize a.franchi@utwente.nl}}
\thanks{$^\dagger$ Department of Computer, Control and Management Engineering, Sapienza University of Rome, 00185 Rome, Italy, {\tt\footnotesize antonio.franchi@uniroma1.it}}
\thanks{This work was partially funded by the Horizon Europe research and
innovation programs under agreement no. 101059875 (Flyflic) and agreement no. 101120732 (Autoassess).}}
\begin{document}
\input{symbols}
\maketitle
\begin{abstract} 
 This work answers positively the question whether non-stop flights are possible for maintaining constant the pose of cable-suspended objects. Such a counterintuitive answer paves the way for a paradigm shift where energetically efficient fixed-wing flying carriers can replace the inefficient multirotor carriers that have been used so far in precise cooperative cable-suspended aerial manipulation.  
 First, we show that one or two flying carriers alone cannot perform non-stop flights while maintaining a constant pose of the suspended object. Instead, we prove that \emph{three} flying carriers can achieve this task provided that the orientation of the load at the equilibrium is such that the components of the cable forces that balance the external force (typically gravity) do not belong to the plane of the cable anchoring points on the load. Numerical tests are presented in support of the analytical results.
\end{abstract}
\section{Introduction}\label{sec:intro}
Aerial robotic object manipulation has been largely studied also thanks to its interesting applications, e.g.,  transportation and assembly, to name a few~\cite{ruggiero2018aerial}. 
Among the several manipulator designs proposed in the literature~\cite{ollero2021past}, 
simple lightweight and low-cost cables have attracted a lot of attention~\cite{drones8020035}, especially due to the flying vehicle payload limitations. 

Uncrewed helicopters have been used in~\cite{bernard2011autonomous} for transportation of suspended objects, while multirotor UAVs (Uncrewed Aerial Vehicles) have been typically considered~\cite{michael2011cooperative, fink2011planning, li2021cooperative,gassner2017dynamic,wahba2024efficient, goodman2022geometric}. In the literature, we find examples of a slung load suspended below a single multi-rotor~\cite{sreenath2013trajectory, pereira2016slung} or a small helicopter~\cite{bernard2009generic}. Two multirotors have been widely exploited, especially for the manipulation of bar-shaped objects~\cite{gabellieri2023equilibria,pereira2018asymmetric,8995928}. Three is the minimum number of robots to control the full pose of a cable-suspended rigid body object~\cite{jiang2012inverse}, and three multirotors have been studied, e.g., in~\cite{9508879,michael2011cooperative, li2021cooperative,masone2016}.


Compared to multirotor UAVs, fixed-wing UAVs have a longer flight endurance, as it has been shown, e.g., in~\cite{leutenegger2016flying}. However, fixed-wing UAVs cannot stop in mid-air, thus making seemingly impossible to keep the load in a static equilibrium while using such platforms as flying carriers. Furthermore, these platforms are unable to Vertical Take Off and Landing (VTOL).
With an intermediate flight endurance, convertible UAVs such as tail sitters have been proposed~\cite{leutenegger2016flying}. These are VTOL platforms that convert themselves to proceed flying as a fixed-wing platform. However, also these designs are obliged to follow non-stop trajectories to ensure an efficient flight. 

Therefore, while non-stop UAVs may provide better endurance for cooperative cable suspended transportation tasks than multirotor UAVs, to the best of the authors' knowledge, it has not been systematically studied yet whether non-stop flights are compatible with keeping the load in a static equilibrium pose, i.e., whether it is possible to have at the same time the pose of the load being static while the flying carriers keep loitering above the object. 

\begin{figure}[t]
    \centering
\includegraphics[trim={0cm 2cm 0cm 2cm},clip,width=0.85\columnwidth]{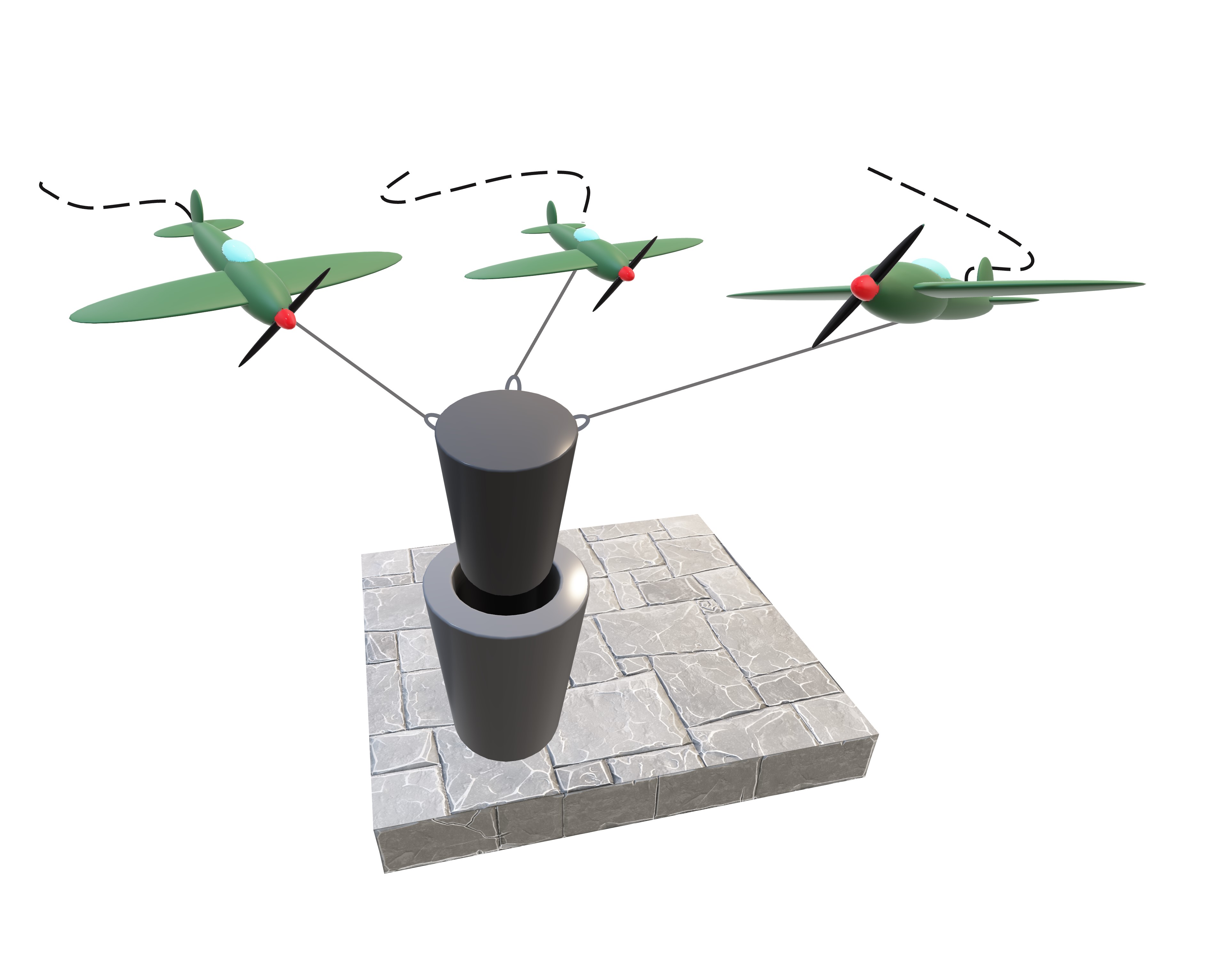}
    \caption{Abstract representation of the concept whose theoretical feasibility is studied in this work. A team of non-stop flying carriers regulate the pose of a suspended object to a static forced equilibrium in an ideal application scenario.}
    \label{fig:fig1}
\end{figure}

The theoretical discovery of such compatibility would pave the way to a completely new type of cable-suspended aerial systems, which would enable a much more energy-efficient execution of aerial manipulation tasks:  long-distance transport would be combined with fine manipulation of the object's pose at the destination or in the presence of difficult intermediate passages, e.g., in cluttered environments. Figure~\ref{fig:fig1} shows an abstract representation of the possible future scenario that would be enabled by this study. 

In this work, we consider the challenges posed by the described intriguing problem and we study analytically whether there exist non-stop trajectories for the flying carriers that are compatible with the constraint of keeping the load at a constant static pose. We focus on the case in which the load must maintain a constant pose because it is obviously the most challenging case compared to the one in which also the load moves. 
The main contributions of the work are as follows: (i) we formalize the problem in mathematical terms; (ii) we prove that with one and two carriers, non-stop flights are not achievable in the considered manipulation scenario; (iii) we prove that \emph{three} is the minimum number of flying carriers for which non-stop flights are possible and (iv) we derive the conditions for that to hold, highlighting degenerate cases in which non-stop flights are prevented; (v) we support the theoretical results through numerical simulation.

The work is organized as follows. First, in Sec.~\ref{sec:model}, we introduce the notation and the problem. Hence, we answer the above-described research question in Sec.~\ref{sec:method}.  In Sec.~\ref{sec:sim}, we show numerical examples. Eventually, conclusions are drawn in Sec.~\ref{sec:conclusion} with an outline of future work. 
\section{Problem Statement}\label{sec:model}
\subsection{Mathematical background}
Consider ${\frameW=\{\originW,\xW, \yW, \zW\},}$  being a world-fixed frame where $\originW$ is the origin, and $\xW, \yW, \zW$ are the x-, y-, and z- axes, respectively. 
Similarly, consider a frame attached to the Center of Mass (CoM) of a rigid body, which represents the manipulated object, and denote  it with ${\frameB=\{\originB, \xB, \yB, \zB\}.}$ The position of $\originB$ in $\frameW$ is denoted with ${\pL\in \nR{3}}$ and $\rotMat_L\in SO(3)$ expresses the attitude of $\frameB$ w.r.t. $\frameW$. The body angular velocity expressed in the body frame is indicated as $^B\angVel_L$.

The rigid body is manipulated through $n$ cables, each one of length $L_i>0$ (with $i=1\ldots,n$) and attached to point $B_i$ on the object, with $^B\vect{b}_i\in\nR{3}$ representing the constant position of $B_i$ expressed in $\frameB$\footnote{The left superscript, $W$ if omitted, expresses the reference frame.}.
Let $\vect{q}_i\in S^2$ be the configuration of the $\ith{i}$ cable in $\frameW$, $T_i$ its tension,  
and 
\begin{equation}
    \vect{f}_i=T_i\vect{q}_i\label{eq:fi}
\end{equation} 
be the coordinates in $\frameW$ of the force that the $\ith{i}$ cable exerts on the load. The cables' mass and inertia are considered negligible compared to the robots' and object's ones, and cables are considered in tension under the effect of the weight~\cite{sanalitro2020full, gabellieri2023equilibria} (and so $T_i\neq0)$. Each cable is attached, at the other end, to a point on an aerial robot (flying carrier), whose position in $\frameW$ is denoted with $\pR{i}\in\nR{3}$.


The dynamics of the transported object is described by
\begin{align}
    \massL\ddpL &=-\massL g \vE{3} + \textstyle\sum_{i=1}^n{\vect{f}_i}\label{eq:load_dyn_tr}\\
    \inertiaL{^B\angAcc}_L &= -\inertiaL{^B\angVel}_L\times {^B\angVel}_{L} + \textstyle\sum_{i=1}^n{S(^B\vect{b}_i) \rotMatL\vect{f}_i}\label{eq:load_dyn_rot}\\
    \drotMatL &= S(\angVel_L)\rotMatL
\end{align}
where $\massL, \inertiaL$ are the mass and the rotational inertia of the rigid body load and $S(\star)$ indicates the skew-symmetric operator  implementing the cross product between two vectors. 

The position and velocity of the $\ith{i}$ flying carrier in  $\frameW$ are, from the system's kinematics, equal to
\begin{align}
\pR{i}&=\pL+\rotMatL{^B\vect{b}}_i+\vect{q}_iL_i\label{eq:kinematics}\\
\dpR{i}&=\dpL+\drotMatL{^B\vect{b}}_i+\dot{\vect{q}}_iL_i.\label{eq:diff_kin}
\end{align}

To compactly rewrite the load dynamics we use the matrix $\matr{G}\in\nR{6\times3n}$ that maps the cable forces to the wrench applied at the object's center of mass, referred to as grasp matrix in the literature~\cite{yoshikawa1999virtual, tognon2018aerial}  
\begin{equation}
    \matr{G}=\begin{bmatrix}
        \eye{3} & \eye{3} & ... & \ \eye{3}\\ S(^B\vect{b}_1)\rotMatL& S(^B\vect{b}_2)\rotMatL&...&S(^B\vect{b}_n)\rotMatL,\label{eq:grasp}
    \end{bmatrix}.
\end{equation}  

The load dynamics~\eqref{eq:load_dyn_tr}-\eqref{eq:load_dyn_rot} can be compactly rewritten as
\begin{align}
    \matr{W}
= \matr{G}\f
\label{eq:compact-dynamics}
\end{align}
where $\f=[\cableForce{1}^\top\ \cableForce{2}^\top\ ...\ \cableForce{n}^\top]^\top$ stacks all forces that the cables apply to the object and ${\matr{W}=\begin{bmatrix}
         \massL(\ddpL+ g \vE{3})  
        \\
        \inertiaL{^B\angAcc}_L+ \inertiaL{^B\angVel}_L\times {^B\angVel}_{L}
    \end{bmatrix}}$.

Inverting~\eqref{eq:compact-dynamics}, see~\cite{sreenath2013dynamics}, we obtain an expression for all the cable forces $\f$ that are compatible with a certain $\matr{W}$
\begin{equation}    \f=\matr{G}^\dagger\matr{W}+\matr{N}\vect{\lambda},\label{eq:all_forces}
\end{equation}
being $\star^\dagger$ indicating the Moore-Penrose pseudo-inverse, $\matr{N}\in\mathbb{R}^{3n\times m}$ being the nullspace projector of the grasp matrix $\matr{G}$ (with $m$ the dimension of the nullpsace of $\matr{G}$), and $\vect{\lambda}\in \mathbb{R}^m$ a column vector containing $m$ free parameters $\lambda_1,\ldots,\lambda_m$ referred to as the \textit{internal forces}~\cite{yoshikawa1999virtual}.

For each load forced equilibrium or trajectory defined by the load state and the corresponding forcing input $\matr{W}$ we can obtain from~\eqref{eq:all_forces} all the cable forces that are compatible with such load equilibrium/trajectory. These forces are parameterized by $\vect{\lambda}$. In turn, to each cable force corresponds a carrier position obtained by~\eqref{eq:kinematics} substituting $\vect{q}_i=\vect{f}_i/T_i$, i.e., $\vect{q}_i=\vect{f}_i/\|\vect{f}_i\|$. So we can conclude that for any given load equilibrium/trajectory there are $\infty^m$ compatible multi-UAV coordinated trajectories which are parameterized by the internal forces $\vect{\lambda}$ and are obtained by plugging~\eqref{eq:all_forces} in~\eqref{eq:kinematics}.


\subsection{Maintaining Load Static Equilibrium with non-stop flights}

Internal forces have been exploited in the literature of cooperative aerial manipulation of suspended objects to, e.g., optimize the tension distribution in the carriers' cables~\cite{masone2016}, perform obstacle avoidance~\cite{li2022safety}, or stabilize the object's pose in a force-based manipulation scenario~\cite{gabellieri2023equilibria}. However, all the aforementioned works consider multirotor UAVs that can stop (i.e., set $\dpR{i}=\vect{0}$) during the manipulation task execution. 

Departing from the previous literature, in this work we explore whether the internal forces can be exploited to allow the flying carriers to keep loitering with $\|\dpR{i}\|\geq \underline{v}>0$ while the load is kept to a static pose i.e., $\dpL={^B\angVel}_{L}=\vect{0}$.

Consider a static equilibrium of the suspended load characterized by position $\pLEq$ and orientation $\rotMatLEq$.   
When the load is at static equilibrium, the only way the cable forces can still change is thanks to a change of the internal forces $\vect{\lambda}$ since they are the only time-varying parameters in~\eqref{eq:all_forces} in that case. For analyzing the effect of changing internal forces we write them as the state of the following dynamical system: \begin{equation*}
    \begin{cases}
\dot{\vect{\lambda}}=\vect{u}_\lambda\\ \vect{\lambda}(0)=\vect{\lambda}_0\end{cases}\end{equation*} where $\vect{u}_\lambda$ acts as an input that we can assign and $\vect{\lambda}_0$ is their initial condition.
\begin{problem}[pose regulation with non-stop flights]\label{prob:nonstop}
Assume that $\pL(0)=\pLEq$ and $\rotMatL(0)=\rotMatLEq$.
We ask if there exists a $n$ and a trajectory $\vect{\lambda}(t)$ such that $\forall t\geq 0,$ \begin{align}
\begin{cases} 
\pL(t)=\pLEq,\\ 
\rotMatL(t)=\rotMatLEq, \\
\|\dpR{i}(t)\|\geq \underline{v}>0, \quad \forall i=1,\ldots,n\\
-\infty<\underline{\lambda}\leq \lambda_i(t)\leq\overline{\lambda}<\infty,\quad \forall i=1,\ldots,n
\label{eq:equilib}
\end{cases}
\end{align} 
\end{problem} 
\smallskip

The first two equations in~\eqref{eq:equilib} impose static equilibrium of the load. The third equation, where $\underline{v}$ is a constant, is a lower bound to the norms of the flying carriers' velocities to impose that they do not stop while maintaining the  desired constant pose of the object. Finally, the last expression in~\eqref{eq:equilib} imposes that each component of $\vect{\lambda}$ is bounded by certain upper and lower bounds, $\overline{\lambda}, \underline{\lambda}$. This last constraint has clear practical reasons as internal forces of too large magnitude in~\eqref{eq:all_forces} generate cable forces that will eventually break the system.


When~\eqref{eq:equilib} holds,~\eqref{eq:all_forces} can be rewritten component-wise as \begin{equation}
    \vect{f}_i(t)=\vect{f}_{0i}+\tilde{\vect{f}}_i(t), \label{eq:each_force}
\end{equation}
where $\vect{f}_{0i}$ is the component of each force that comes from the first term in the left-hand side (LHS) of \eqref{eq:all_forces}, namely $\matr{G}^\dagger\matr{W}$; it balances the external wrench and is constant when the load is static. Instead, $\vect{\tilde{f}}_i$ is the variable component of the cable force generated by the internal forces, which comes from $\matr{N}\vect{\lambda}(t)$  in \eqref{eq:all_forces}.

Taking the time derivative of~\eqref{eq:all_forces} we obtain:
\begin{equation}
\dot{\vect{f}}={\matr{N}}\dot{\vect{\lambda}}.
\label{eq:all_forces_deriv}
\end{equation}
Equation~\eqref{eq:all_forces_deriv} gives us a relationship between the internal force variation and the total cable force variation; to give an intuition, it gives all the possible variations of the cable forces, and hence of the carriers' velocities, that do not perturb the object pose equilibrium.

By differentiating \eqref{eq:fi} w.r.t. time, the cable force variation~\eqref{eq:all_forces_deriv} it holds
\begin{equation}
\dot{\vect{f}}_i=\dot{T}_i\vect{q}_i+T_i\dot{\vect{q}}_i,\label{eq:df}
\end{equation}
i.e., the force variation is composed of two orthogonal components: one is directed as $\vect{q}_i$ (along the cable), and one as $\dot{\vect{q}}_i$ (orthogonal to the cable).
\section{Non-stop Flight Analysis}\label{sec:method}
In this section, we will reply to the question: `are non-stop carrier UAV flights compatible with keeping  cable-suspended objects at a static pose?', which is formally stated in Problem~\ref{prob:nonstop}.

First, we introduce here a few tools to reply to the above-formulated research question. We start by formalizing an intuitive condition. 
\begin{prop}\label{prop_1}
When the load is at a static equilibrium and the cables are in tension,    
$$\exists \underline{v}>0 \text{~s.t.~}  \forall  t \;\|\dpR{i}(t)\|  \geq \underline{v}
 \Longrightarrow 
\exists s>0 \text{~s.t.~} \forall  t \;\|\dot{\vect{f}}_i(t)\|\geq s $$
\end{prop}
\begin{proof}
By computing~\eqref{eq:diff_kin} at the load's static equilibrium, we have 
\begin{equation}
    \dpR{i}=\dot{\vect{q}}_i L_i.\label{eq:dp_eq_dqL}
\end{equation} 
    Hence, by using $\|\dpR{i}\|\geq \underline{v}$ inside~\eqref{eq:dp_eq_dqL} and given  that the cable length $L_i$ is greater than zero, we obtain \begin{equation}
        \|\dot{\vect{q}}_i\|\geq\frac{\underline{v}}{L_i}.\label{eq:bounded_dq}\end{equation}
        Consider now that, from~\eqref{eq:df}, the cable forces vary along two orthogonal directions. That, plus the fact that  $\vect{q}_i$ is a unit vector by definition, leads to \begin{equation}
    \|\dot{\vect{f}}_i\|=|\dot{T}_i| + |T_i|\|\dot{\vect{q}}_i\|\label{eq:norm_df}
        \end{equation}
        Now, starting from~\eqref{eq:df} and using~\eqref{eq:bounded_dq},~\eqref{eq:norm_df}, $L_i>0$ and $|T_i|\neq0$ by hypothesis of tensioned cables,  we write:
    $$\|\dot{\vect{f}}_i\|=|\dot{T}_i| + |T_i|\|\dot{\vect{q}}_i\|\geq|\dot{T}_i|+|T_i|\frac{\underline{v}}{L_i}\geq|T_i|\frac{\underline{v}}{L_i}:=s>0.$$
\end{proof} 
\begin{remark}
Proposition~\ref{prop_1} tells us that the  lower boundedness of the norms of the cable forces derivatives is a necessary condition for~\eqref{eq:equilib} to hold. Namely, in reverse, if the norms of the cable forces derivatives are not lower bounded, then non-stop flights are impossible (i.e., $\nexists s>0\text{~s.t.~}  \forall  t\ \|\dot{\vect{f}}_i(t)\|\geq s\Rightarrow \nexists \underline{v}>0\text{~s.t.~}  \forall  t\  \|\dpR{i}\|\geq \underline{v}$). 
\end{remark}


\subsection{1-carrier case}

First, we consider the case in which the is only one flying carrier, i.e., it is $n=1$. The sole equilibrium points of the system are in this case the ones in which the load's center of mass lies below the vertical cable that suspends it\footnote{Despite this number of carrier(s) does not allow regulating the full pose of a suspended object, we are here interested in understanding if the carrier(s) can loiter while the object is at a static equilibrium.\label{footnote:not-controlled-pose}}.

The grasp matrix in~\eqref{eq:grasp} is 
$$ \matr{G}=\begin{bmatrix}
        \eye{3}\\ S(^B\vect{b}_1)\rotMatL
    \end{bmatrix}. $$ Hence, as the grasp matrix $\matr{G}\in\nR{6\times3}$ has number of columns smaller than the number of rows, it has a trivial nullspace and $\matr{N}=\matr{0}$. \begin{fact}\label{fact:1robot}
        For $n=1$, $\nexists t$ such that conditions~\eqref{eq:equilib} hold.
    \end{fact}
    \begin{proof}
    $\matr{N}=\vect{0}$ implies that~\eqref{eq:all_forces_deriv} becomes $\dot{\vect{f}}\equiv\vect{0}$. 
    According to~\eqref{eq:df}, the cable force varies along two orthogonal components. Hence, the only way in which $\dot{\vect{f}}_i=\vect{0}$ is that both those orthogonal components are equal to zero. In other words, both $\dot T_i=0$ and $\dot{\vect{q}}_i=\vect{0}$ (because $T_i\neq 0$ by assumption). 
    When the load is at equilibrium, $\dpL=\vect{0}$ and $\drotMatL=\vect{0}$. By applying all the above conditions to~\eqref{eq:diff_kin}, one obtains $\dpR{i}=\vect{0}$.
\end{proof} 
In conclusion, Fact~\ref{fact:1robot} tells that for $n=1$ non-stop flights for maintaining the static equilibrium of a suspended object \textit{are not possible}.

\subsection{2-carrier case}
For $n=2$ flying carriers, the load grasp matrix is$^{\ref{footnote:not-controlled-pose}}$  $$\matr{G}=\begin{bmatrix}
        \eye{3} & \eye{3} &\\ S(^B\vect{b}_1)\rotMatL& S(^B\vect{b}_2)\rotMatL&
    \end{bmatrix}.$$  $\matr{G}\in\nR{6\times6}$ has a non-trivial nullspace of dimension $m=1$ and a basis of it is built from the vector along the direction connecting the cable attachment points on the load, $B_1$ and $B_2$~\cite{gabellieri2023equilibria}. 

In particular, $\matr{N}={\begin{bmatrix}\vect{b}_{1,2}^\top&\vect{b}_{2,1}^\top\end{bmatrix}^\top},$ 
 where we indicate in general the unit vector in the direction $B_iB_j$ as ${\vect{b}_{i,j}=\frac{\vect{b}_i-\vect{b}_j}{|\vect{b}_i-\vect{b}_j|}}$, with $(i,j)\in\{(1,2),(2,1)\}$. 

\begin{fact}\label{fact2}
    For $n=2$, $\nexists \vect{\lambda}(t)$ such that $\forall t$~\eqref{eq:equilib} holds.
\end{fact}

\begin{proof}
We proceed by contradiction. At the load equilibrium, $\vect{b}_{i,j}$ is constant by construction. Equation~\eqref{eq:all_forces_deriv} leads to
 \begin{equation}
     \dot{\vect{f}}_i= \dot{\lambda}_1{\vect{b}_{i,j}}\label{eq:df_2robots}.
 \end{equation}
Assume that \eqref{eq:equilib} holds. Then, for Proposition \ref{prop_1},   $\exists s>0: \|\dot{\vect{f}}_i\|\geq s$.  For the considered case of 2 carriers,  $\forall t>0,$  $\exists s>0:\|\dot{\vect{f}}_i\|\geq s\Longleftrightarrow|\dot{\lambda}_1|\geq s.$ This follows from~\eqref{eq:df_2robots} considering that $\vect{b}_{i,j}$ has unit norm by construction. 
Considering that $|\dot{\lambda}_1|\geq s$ must hold $\forall t$ and that $\dot{\lambda}_1$ is a continuous function of $t$, then either $\dot{\lambda}_1\geq s$ $\forall t$ and thus $\lambda_1$ is strictly  
increasing, or  $\dot{\lambda}_1\leq -s$ $\forall t$  
and thus $\lambda_1$ is strictly 
decreasing. In both cases, $\lambda_1$
 is unbounded for 
$t\to\infty$.  
This contradicts the boundedness of $\lambda_1$ imposed in~\eqref{eq:equilib} and therefore~\eqref{eq:equilib} cannot hold as two of its conditions are in conflict.
\end{proof} 



In conclusion, Fact~\ref{fact2} tells us that non-stop UAV flights for the pose regulation of a cable-suspended object manipulated by 2 carriers \textit{are not possible}.  

\subsection{3-carrier case}
 In this section, we consider the case in which $n=3$ carriers are employed. Let us restrict the analysis to the non-degenerate cases in which the cable attachment points on the objects, the $B_i$'s, are not all aligned with each other. In that degenerate case, the problem is similar to the 2-carrier scenario.  In our study, we are going to assume that any two vectors $\vect{b}_{i,j}$ and $\vect{b}_{i,k}$  with $(i,j,k)\in\{(1,2,3),(2,3,1),(3,1,2)\}$ are sufficiently misaligned. In particular, there exists $\theta_{\text{min}}>0$ that lower bounds any possible angle generated by the lines that the pairs $\vect{b}_{i,j}$, $\vect{b}_{i,k}$ belong to.  

For $n=3$, $\vect{\lambda}=[\lambda_1 \,\lambda_2 \, \lambda_3]^\top$, and a choice for $\matr{N}$ is expressed starting from the directions connecting the three points $B_i$ in pairs.  In particular, it holds that~\cite{sreenath2013dynamics, yoshikawa1999virtual}:
  \begin{equation}
      \matr{N}=\begin{bmatrix}
          \vect{b}_{1,2} & \vect{0} & \vect{b}_{1,3}\\
          -\vect{b}_{1,2} & \vect{b}_{2,3} & \vect{0}\\
          \vect{0} & -\vect{b}_{2,3} & -\vect{b}_{1,3}
      \end{bmatrix}\label{eq:3_N}.
  \end{equation}
  Equation~\eqref{eq:3_N} tells us that the components of each cable force generated by the internal forces are a couple of equal and opposite forces acting along the line connecting two cable attachment points. 
In the following, we show that the necessary condition of Proposition 1 becomes sufficient for $n=3$ under certain assumptions.
\begin{figure}[!h]
    \centering
    \includegraphics[width=0.85\columnwidth]{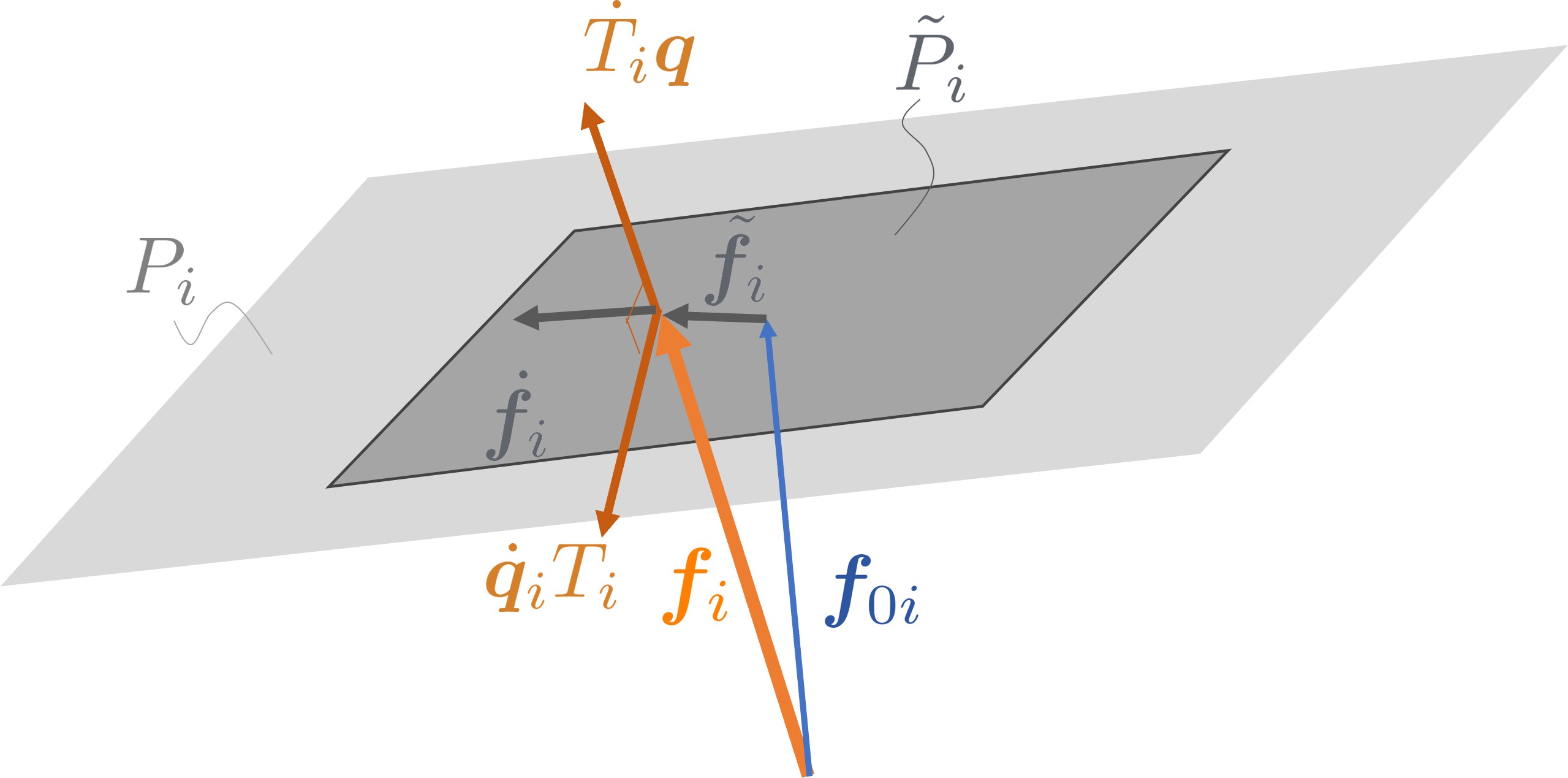}
    \caption{Geometric representation of the cable force, its components, and its derivative. The grey vector $\dot{\vect{f}}_i$ belongs to the plane $P_i$, while its components $T_i\dot{\vect{q}}_i, \dot{T}_i \vect{q}_i \notin P_i$ (in general). $\tilde{\vect{f}}_i$ is contained in the subset $\tilde{P}_i$. }
    \label{fig:planeP}
\end{figure}
\begin{figure}[!h]
    \centering
    \includegraphics[width=0.65\columnwidth]{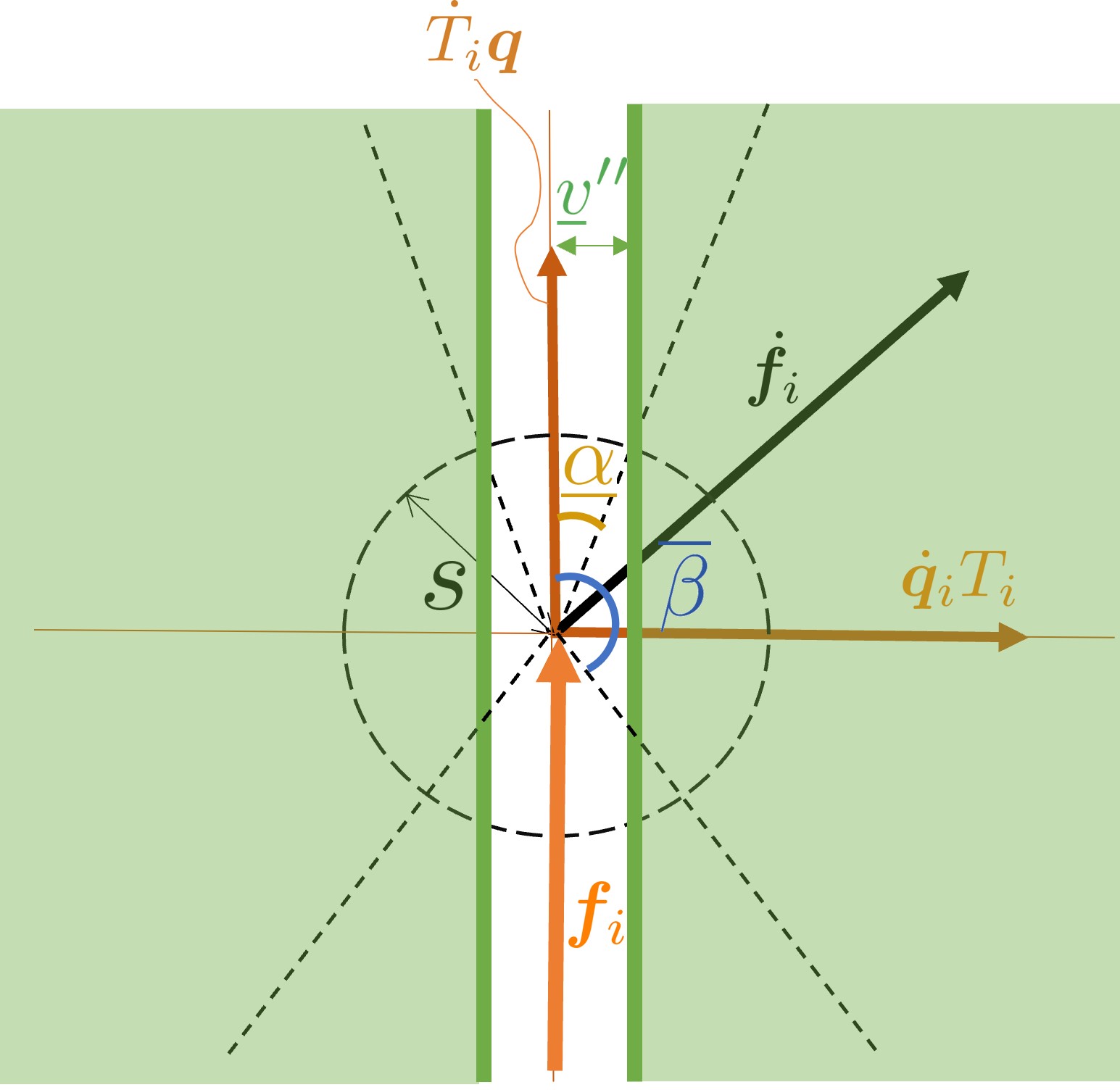}
    \caption{Representation of the plane including $\dot{\vect{f}}_i$ and its components along the directions parallel and perpendicular to the corresponding cable force. The minimum and maximum values of the angle between $\dot{\vect{f}}_i$ and $\vect{f}_i$ are represented with dotted half-lines, incident in the middle point, and the circle of radius $s$ is the lower bound on the norm of $\dot{\vect{f}}_i$. As a consequence, the norm of the component of $\dot{\vect{f}}_i$ on the direction orthogonal to the cable force, namely along $\dot{\vect{q}}_i$, is lower bounded by a positive quantity $\underline{v}''$. }
    \label{fig:plane_bounds}
\end{figure}

 \begin{figure*}[t]
    \centering
    \includegraphics[width=0.34\textwidth,trim={3cm 9cm 3cm 11cm},clip]{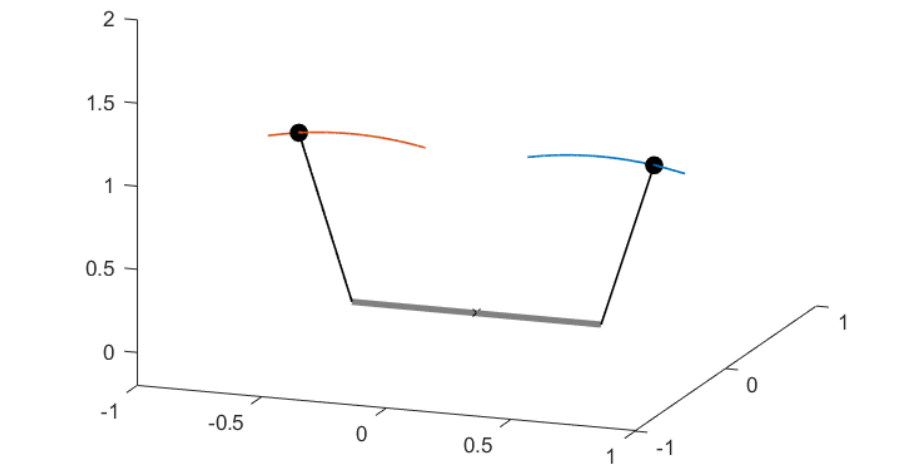}
  \qquad\includegraphics[width=0.25\textwidth,trim={3cm 9cm 3cm 9cm},clip]{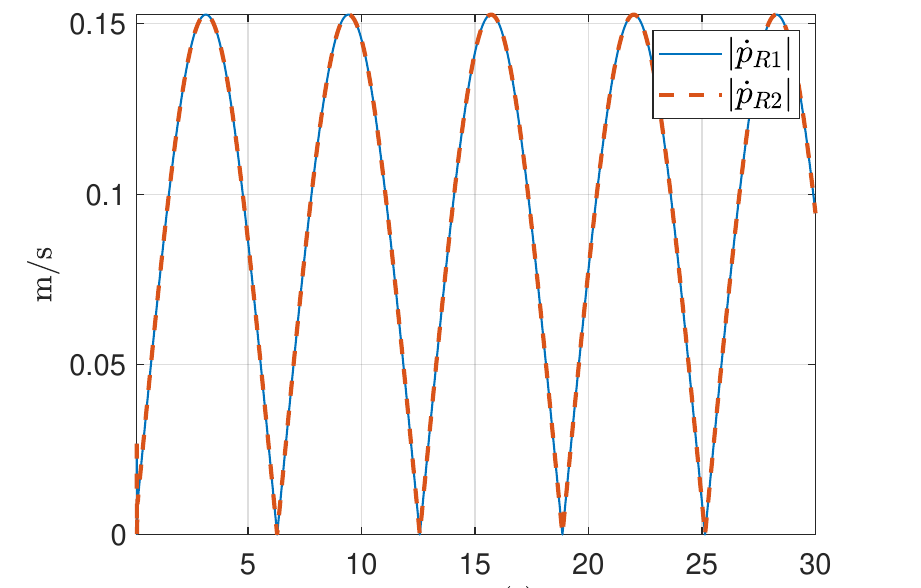}
  \qquad\includegraphics[width=0.25\textwidth,trim={3cm 9cm 3cm 9cm},clip]{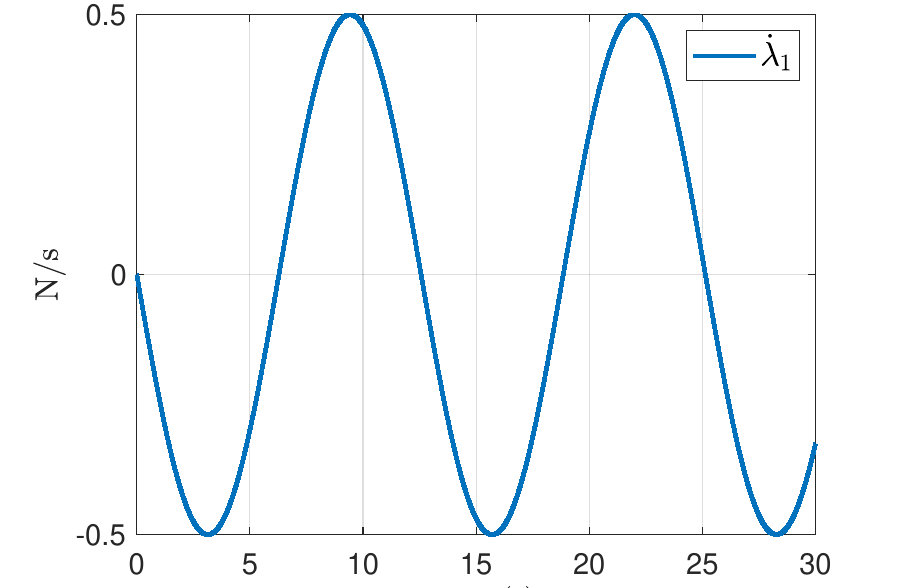}
    \caption{On the left: two carriers (black dots) cannot maintain the pose of the object (in grey) while performing non-stop flights; the cables are in black, and colored curves are the carrier's paths. Carriers' velocities and $\dot{\vect{\lambda}}$ are reported in the other two plots. }
    \label{n2}
\end{figure*}
\begin{prop}\label{prop2} \label{prop:f_lower_bound}
    For $n=3$, if the load's attitude at the static equilibrium is such that $\forall i$   $\vect{f}_{0i}\notin{\rm{span}}\{ \vect{b}_{i,j},\vect{b}_{i,k} \}$, with
    $i,j,k\in\{1,2,3\},$ $i\neq j\neq k \neq i$ 
then the following is true
 ${\forall i=1,2,3}$    
    $${\exists s>0 \text{ s.t }} \forall t\ \|{\dot{\vect{f}}_i(t)\|\geq s 
\; \Rightarrow \;
\exists \underline{v}>0 \text{ s.t } \forall t\ 
\|\dpR{i}(t)\|\geq\underline{v}}$$ 
\end{prop}
    
\begin{proof} 
First of all, from \eqref{eq:diff_kin} at the equilibrium, ${\|\dpR{i}\|=L_i\|\dot{\vect{q}}_i\|}$; hence, to prove the thesis of the proposition it is equivalent to show that $\|\dot{\vect{q}}_i\|\geq \underline{v}'$ for a certain $\underline{v}'>0$. 
For the sake of a streamlined presentation, we avoid presenting all the mathematical details and to make use of explanatory figures that help the intuition of the reader being convinced about the validity of the proof. 

Please refer to Fig.~\ref{fig:planeP} to better understand the quantities defined in the following. 
Referring to~\eqref{eq:each_force} we know that ${\vect{f}_i=\vect{f}_{0i}+\tilde{\vect{f}}_i}$, where $\vect{f}_{0i}$ 
is obtained by the corresponding entries of $\matr{G}^\dagger\vect{W}$ and it is nonzero and constant in time, and $\tilde{\vect{f}}_i$ is obtained by the corresponding entries of $\matr{N}\vect{\lambda}$ and it is
time-varying and belongs to the plane $P_i$ spanned by the vectors $\vect{b}_{i,j}$ and $\vect{b}_{i,k}$ with $j,k\neq i$ and $j\neq k$, according to~\eqref{eq:3_N}. 
From the hypotheses of the proposition, we have that $\vect{f}_{0i}$ does not belong to $P_i$. Since $\vect{f}_{0i}$ is constant and nonzero, also $\vect{f}_i$ does not belong to $P_i$.
On the other hand, from~\eqref{eq:all_forces_deriv}, we have that $\dot{\vect{f}}_i$ belongs to $P_i$. 
As a consequence, thanks also to the hypothesis that  $\dot{\vect{f}}_i\neq \vect{0}$, we obtain that $\dot{\vect{f}}_i$ and $\vect{f}_i$ are never parallel to each other (two parallel vectors must belong to the same plane). In other words, the angle between each cable force and its
derivative is $0<\angle (\vect{f}_i, \dot{\vect{f}}_i)<\pi.$ 

Because of the boundedness of each component of $\vect{\lambda}$, the endpoint of $\vect{f}_i$ is constrained inside a 
bounded set denoted with $\tilde{P}_i\subset P_i$, the dark grey region in Figure \ref{fig:planeP}. $\tilde{P}_i$ is centered around the end-point of $\vect{f}_{0i}$ and contains the vector $\tilde{\vect{f}}_i.$ 
For each $\vect{f}_i$, with $i=1,2,3$, we have a minimum and maximum of the angle $\angle (\vect{f}_i, \dot{\vect{f}}_i)$ when $\dot{\vect{f}}_i$ varies in $\tilde{P}$, which we denote as 
$\alpha(\vect{f}_i)$ and  $\beta(\vect{f}_i)$
respectively. We denote with $\underline{\alpha}$ and $\overline{\beta}$ the minimum and maximum of $\alpha(\vect{f}_i)$ and  $\beta(\vect{f}_i)$ when the end point of $\vect{f}_i$ varies in $\tilde{P}_i$,  and $i$ varies in $\{1,2,3\}$, respectively. Since those functions are continuous and the set $\tilde{P}_i$ is bounded such minimum and maximum exist and therefore, for what has been previously said, they are such that $0<\underline{\alpha}\leq\overline{\beta}<\pi.$

Now let's consider the two orthogonal components of $\dot{\vect{f}}_i$ defined in~\eqref{eq:df}, i.e. $\dot{T}_i\vect{q}_i $ and $T_i\dot{\vect{q}}_i$. These three vectors define a plane that contains also $\vect{f}_i=T_i\vect{q}_i$.
We represent such a plane in Figure \ref{fig:plane_bounds}, where the bounds of the angles between $\dot{\vect{f}}_i$ and the direction of the force $\vect{f}_i$, given by $\vect{q}_i$, are represented by the dashed lines. Moreover, the lower bound of the norm of the force derivative, $s>0$, is represented by a dotted circle; the endpoint of $\dot{\vect{f}}_i$ is constrained to be outside of that circle. Hence, its component on the direction orthogonal to the force, namely the one parallel to $\dot{\vect{q}}_i$, must lie in the green regions in Figure~\ref{fig:plane_bounds}, delimited by the lines at distance $\underline{v}''>0$ from the line where $\vect{f}_i$ lies. This gives a geometrical proof that $\|\dot{\vect{q}}_i\|T_i \geq \underline{v}''$ and so $\|\dot{\vect{q}}_i\|\geq \frac{\underline{v}''}{T_i}:=\underline{v}'>0.$
\end{proof} 

 We are now ready to state the main result. \begin{fact}\label{fact3}
    If the load's static equilibrium is such that $\forall i$   $\vect{f}_{0i}\notin{\rm{span}}\{ \vect{b}_{i,j},\vect{b}_{i,k} \},$ with $i,j,k\in\{1,2,3\},$ $i\neq j\neq k \neq i$, then for $n=3$ there exists infinite trajectories $\vect{\lambda}(t)$ such that $\eqref{eq:equilib}$ is verified.
    A possible choice is to select $\forall i\in\{1,2,3\}$ the following sinusoidal functions \begin{equation}\label{eq:cos}
\lambda_i(t) = \lambda_i(0)+A\cos{(\psi t+\Phi_i)}
    \end{equation}
    where $\psi>0$; 
    the amplitude $A$ is constant and bounded ($0<A<\infty$); the initial condition is bounded $\underline{\lambda}\leq\lambda_i(0)\leq\overline{\lambda}$; and the constant phases $\Phi_i$ are such that $0<\underline{\Phi_i}\leq\Phi_i\leq \bar{\Phi}<\pi$, and that, for every $j\neq i$  $|\Phi_i-\Phi_j|\geq\phi>0$ (non-coinciding phases).
\end{fact}
\begin{proof}
We are going to show that, choosing the aforementioned trajectories, $\|\dot{\vect{f}}_i\|$ is lower bounded, so we can apply Proposition \ref{prop2}.  Let us consider without loss of generality the force applied by cable 1. From equation~\eqref{eq:all_forces_deriv}, 
 \begin{equation}\label{eq:three-robot-f1}
     \dot{\vect{f}}_1=\dot{\lambda}_1\vect{b}_{1,2} + \dot{\lambda}_3\vect{b}_{1,3}.
 \end{equation}
 When ${\lambda}_1$ and ${\lambda}_3$ are chosen as in~\eqref{eq:cos}, they are bounded because the initial conditions and amplitudes $A$ are bounded. Especially, $\forall t,$ $|\lambda_i(t)|\leq |\lambda_i(0)|+|A|$. Moreover,  their derivatives are $\dot{\lambda}_i=-A\psi\sin{(\psi t+\Phi_i)}$. 
 Since $\vect{b}_{1,2}$ and $\vect{b}_{1,3}$  are not parallel, for $\|\dot{\vect{f}}_1\|$ to be lower bounded it is sufficient to show that at least one among $|\dot{\lambda}_1(t)|$ and $|\dot{\lambda}_3(t)|$ is non-zero at any time $t$, which is ensured by the non-coinciding phase assumption on $\Phi_1$ and $\Phi_2$.
 \end{proof}
\textit{Remark: Fact~\ref{fact3} positively answer Problem~\ref{prob:nonstop} telling us that pose regulation of a suspended object using non-stop flying carriers is possible for $n=3$.}

  \begin{figure*}[t]
   \subfloat[Three flying carriers execute non-stop flights not perturbing the pose of the suspended objetc.]{\includegraphics[width=0.31\textwidth,trim={3cm 9cm 3cm 10cm},clip]{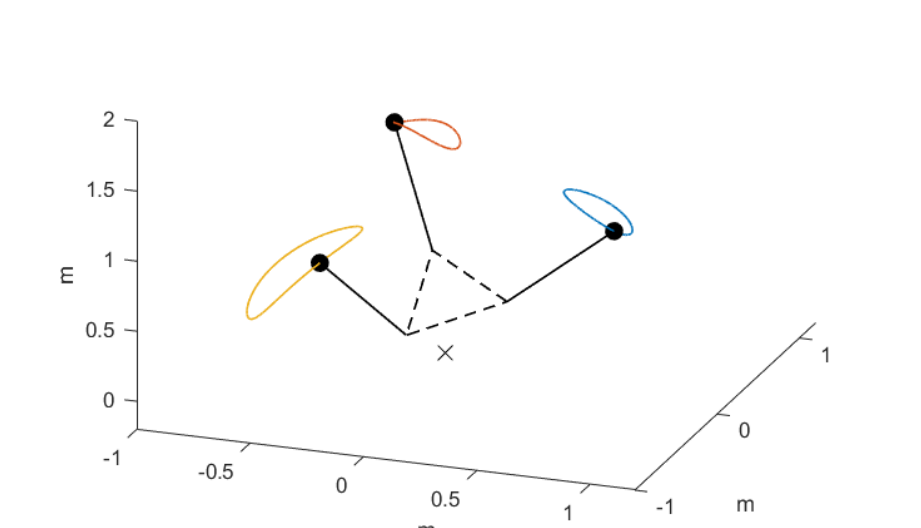}\label{fig:case0}} \quad
   \subfloat[One of the carriers stops when two components $\dot{\lambda}_i$ are equal to each other.]{ \includegraphics[width=0.31\textwidth,trim={3cm 9cm 3cm 10cm},clip]{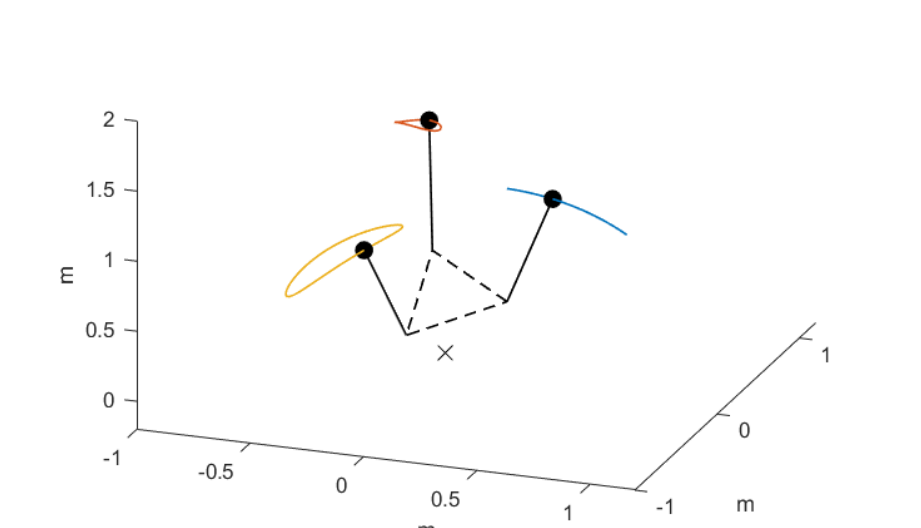}    \label{fig:case1}}\quad
    \subfloat[One of the carriers stopswhen the two components $\dot{\lambda}_i$ vanish simultaneously.]{\includegraphics[width=0.31\textwidth,trim={3cm 9cm 3cm 10cm},clip]{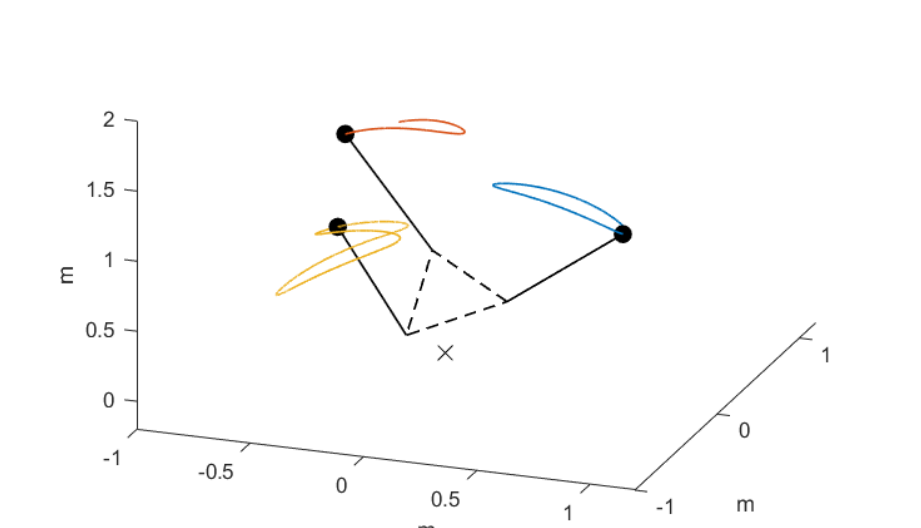}  \label{fig:case2}}\\
    \subfloat[\label{case0_dp}]{\includegraphics[width=0.30\textwidth,trim={3cm 8cm 3cm 9cm},clip]{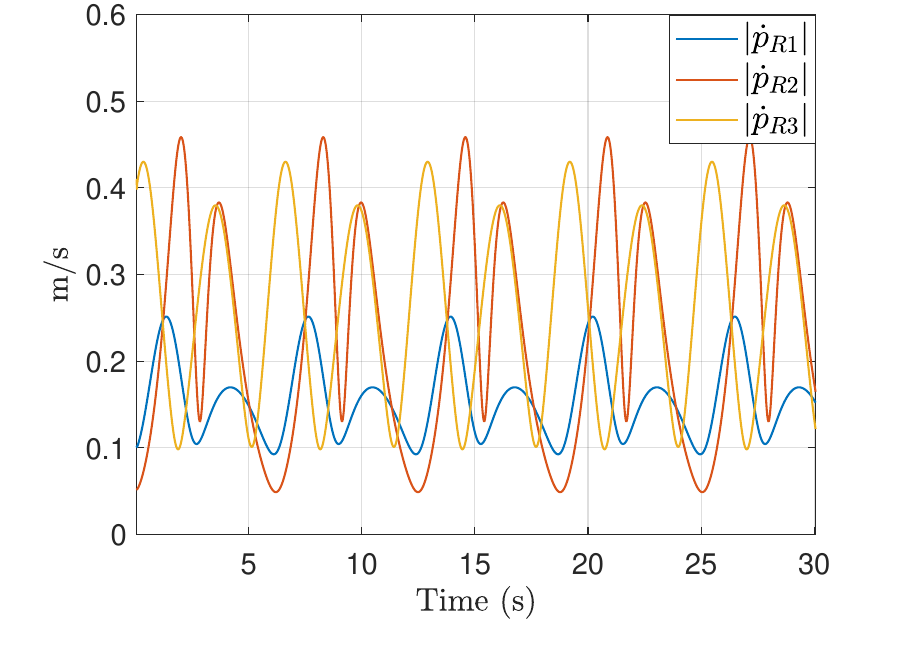}}\qquad  \quad\subfloat[\label{case1_dp}]{\includegraphics[width=0.30\textwidth,trim={3cm 8cm 3cm 9cm},clip]{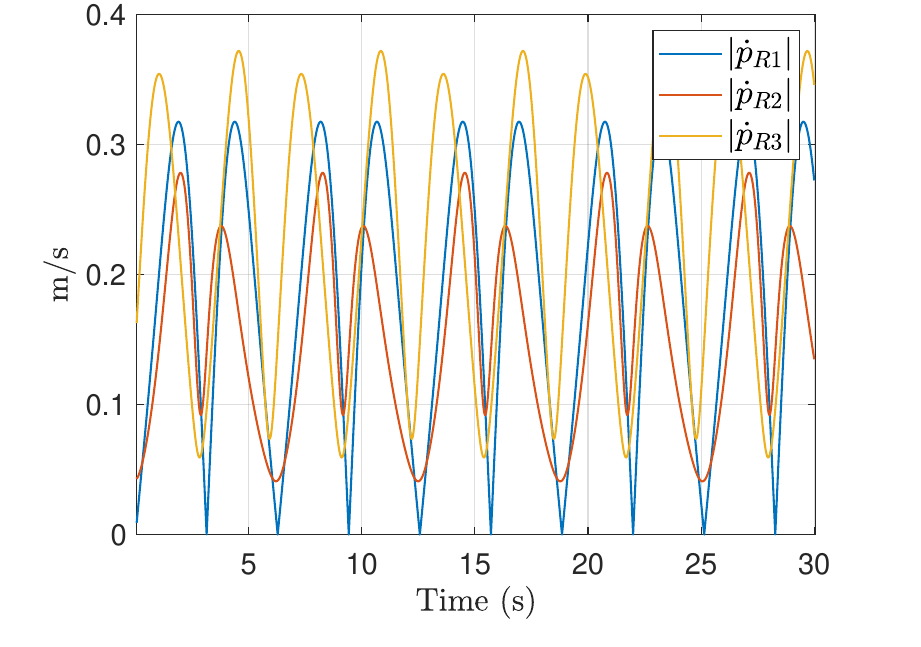}}\qquad  \quad\subfloat[\label{case2_dp}]{\includegraphics[width=0.30\textwidth,trim={3cm 8cm 3cm 9cm},clip]{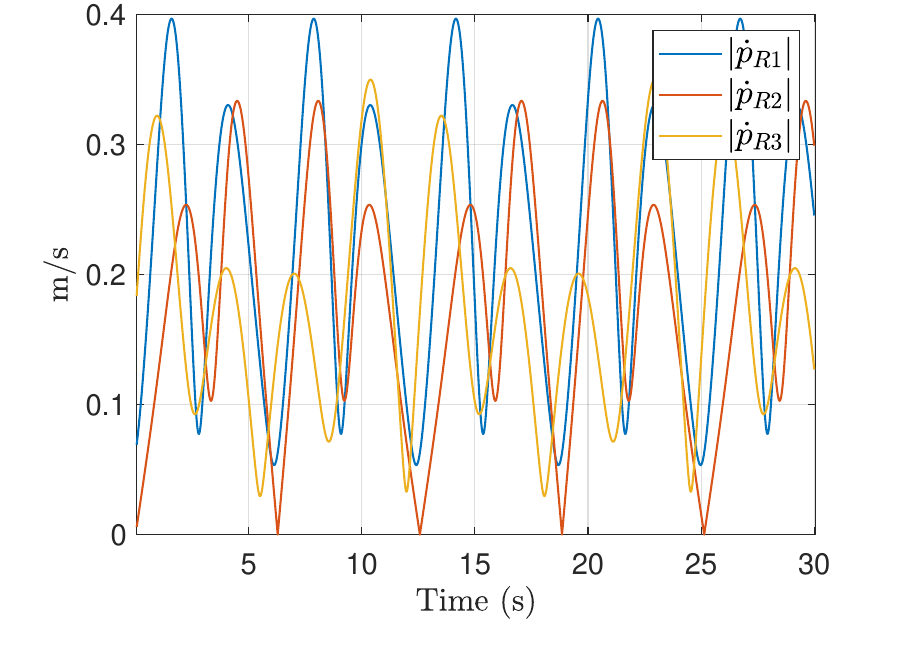}}\\
      \subfloat[\label{case0_dlambda}]{\includegraphics[width=0.30\textwidth,trim={3cm 8cm 3cm 9cm},clip]{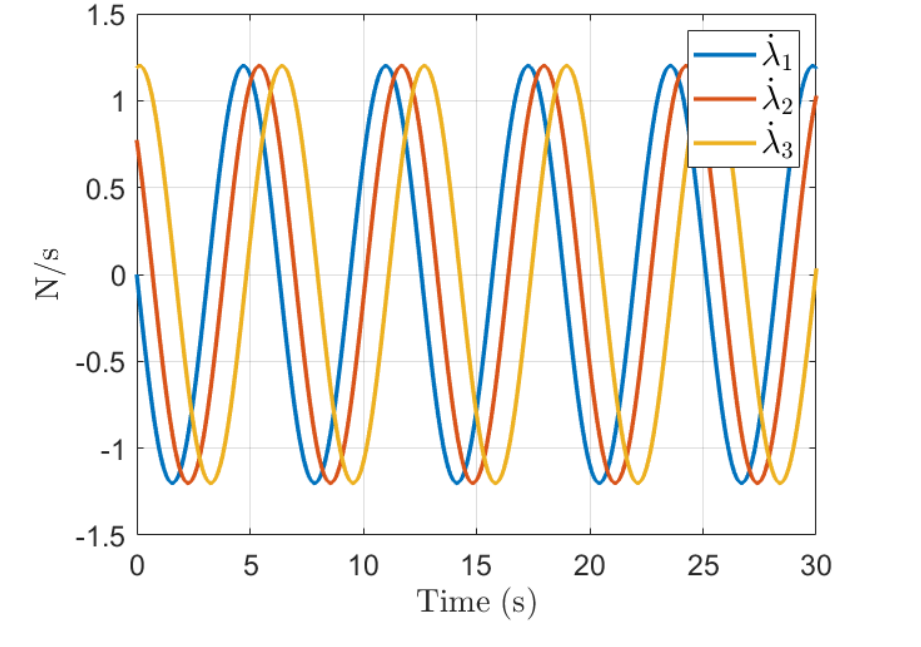}}\qquad  \quad\subfloat[\label{case1_dlambda}]{\includegraphics[width=0.30\textwidth,trim={3cm 8cm 3cm 9cm},clip]{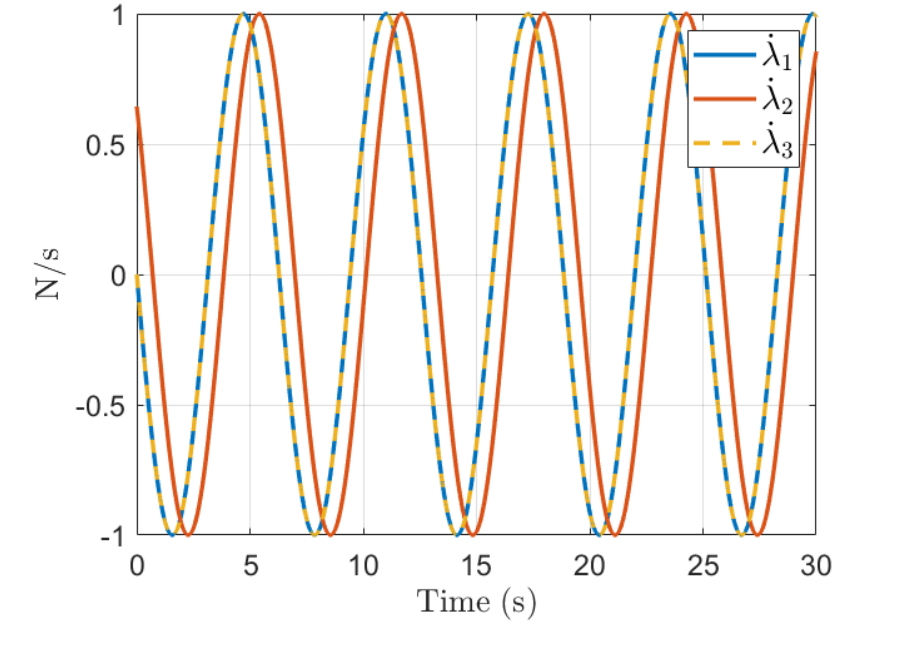}}\qquad  \quad\subfloat[\label{case2_dlambda}]{\includegraphics[width=0.30\textwidth,trim={3cm 8cm 3cm 9cm},clip]{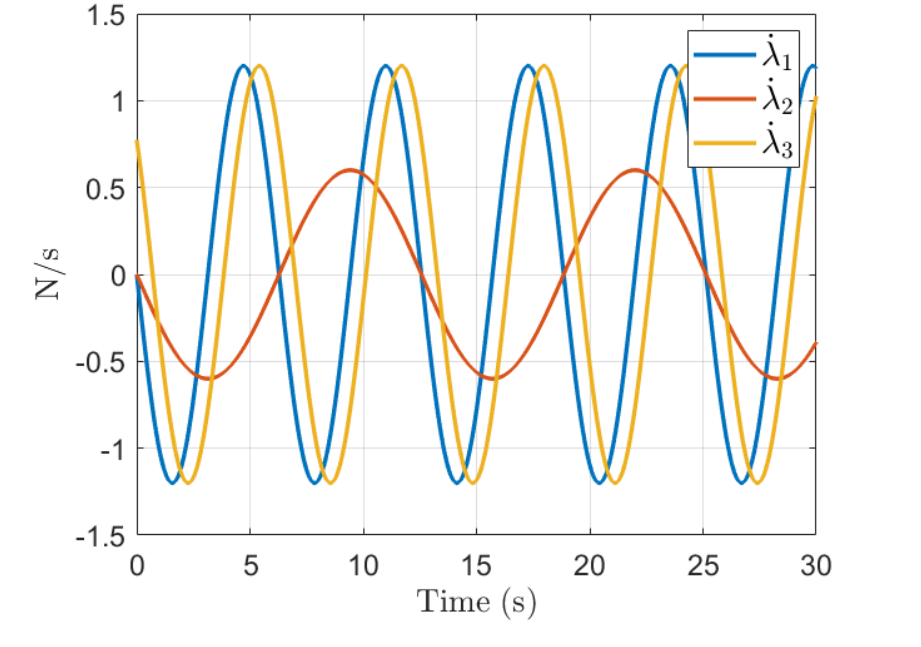}}
     \caption{Results of three different simulations, one in each column. On the leftmost column, non-stop flights are shown; in the other two columns, cases in which the carriers stop. On the first row of each simulation, a frame in which the 3 flying carriers (black dots) maintain the suspended object at a constant pose. Dotted lines connect the cables' attachment points on the object; the cables are black lines; the path of each carrier is a colored line (blue for carrier 1, red for carrier 2, and yellow for carrier 3); the cross is the object CoM. The values of $\|\dpR{i}\|$ and $\dot{{\lambda}_i}$ are also reported for each simulation.}
\end{figure*}
\subsection{Degenerate situations in the 3-carrier case}

Even though we showed that it is possible for 3 flying carriers to maintain a constant pose of suspended objects without stopping, it is also interesting to analyze special choices of $\dot{\vect{\lambda}}$, if any, that do not generate non-stop flights even for 3 carriers. 
 \paragraph{3 carriers\textemdash degenerate case 1}
 If two components of $\dot{\vect{\lambda}}$, let us consider without loss of generality $\dot{\lambda}_1$ and $\dot{\lambda}_3$, are such that $\dot{\lambda}_3=a\dot{\lambda}_1$, $a\in\nR{}$, then the corresponding force variation, in this case as in~\eqref{eq:three-robot-f1}, becomes
 \begin{equation}
     \dot{\vect{f}}_1(t)=(a+1)\dot{\lambda}_1(t)(\vect{b}_{1,2} + \vect{b}_{1,3}),\label{eq:three-robot-f1_a} 
 \end{equation}
  which implies that $\dot{\vect{f}}_1$ lies along a constant direction over time, similar to what happens for the two-carrier case. In fact, since $\vect{b}_{1,2}$ and $\vect{b}_{1,3}$ are constant vectors 
  this case resembles the two-carrier case displayed in~\eqref{eq:df_2robots}. So, for the same reasoning of the proof of Fact~\ref{fact2}, the velocity of carrier 1 is not lower bounded for all time, and so~\eqref{eq:equilib} does not hold. 
 %
%
 \paragraph{3 carriers\textemdash degenerate case 2}
 In any case in which $\exists t:\dot{\lambda}_i(t)=\dot{\lambda}_j(t)=0$ for some $i\neq j\in\{1,2,3\}$ then the corresponding force is zero at that time instant. For the same reasoning done in the proof of Fact~\ref{fact:1robot}, the corresponding carrier stops. Differently from the situation described in case~\emph{a)}, the cable force variation generally is not on a straight line.

\section{Numerical Results}\label{sec:sim}
In this section, we propose simulation results that support the previously presented results. A video of the simulation results can be found at \href{https://youtu.be/A18oq4oTQl8}{https://youtu.be/A18oq4oTQl8}. The simulations have been carried out in Matlab-Simulink. The dynamics of each carrier is a double integrator controlled through a PD feedback law to follow a desired trajectory.  Indeed, the scope of this work is to assess the feasibility of non-stop flights for the considered application. The implementation considering more realistic non-stop UAV dynamics is left for future work. 

The carriers' mass is 0.1 kg, and the derivative and proportional gain matrices are diagonal with values on the diagonals equal to 1.5 Ns/m and 1000 N/m, respectively. 

The rigid body object is suspended below the cables, it has a mass equal to 1 kg and diagonal rotational inertia with diagonal elements equal to 0.01 N$\cdot$m/s. 

Viscous friction on the load and cables' elasticity have been introduced in the simulator to test the theoretical results on more realistic conditions. Viscous friction on the translational and rotational dynamics is equal to 0.1 N$\cdot$s/m and 0.1 rad$\cdot$s/m is added to the object's dynamics to simulate friction with the air. The cables are massless linear springs with a rest length of 0.8 m and rigidity $K_c=500$ $\frac{\rm{N}}{\rm{m}}$. 

Each internal force is chosen as in~\eqref{eq:cos}, with parameters given in the following. The corresponding cable force variations are computed as in~\eqref{eq:all_forces_deriv}  and the carriers' trajectories from~\eqref{eq:kinematics} and~\eqref{eq:diff_kin}.

First, an example with 2 carriers is in Figure~\ref{n2}. The cable attachment points are such that $\vect{b}_{1,2}=\pm[0.5\ 0\ 0]$ m; $\lambda_1(0)=1$ N,  $A_i=1$ N,  $\psi_1=0.5$ rad/s, 
 $\Phi_1=0$. As predicted from the theory,  the carriers stop during their motion when $\dot{\lambda}_1$ change sign, which is ultimately necessary to keep bounded cable forces.

Figure~\ref{fig:case0} shows frames in which 3 carriers manipulate the object with $\forall i,\ \lambda_i(0)=2$ N, $A=1.2$ N, $\psi=2$ rad/s, $\Phi_1=0$, $\Phi_2=0.7$ rad, and $\Phi_3=1.7$ rad. The cable attachment points for 3 carriers are such that $\vect{b}_1=[0.259\    0.034\     0.399]^\top$ m, $
\vect{b}_2=[-0.156 \   0.269 \   0.556]^\top$ m, and $
\vect{b}_3=[-0.1223\  -0.1399  \  0.1778]^\top$ m. As all $\lambda_i$ vary according to~\eqref{eq:three-robot-f1}, at the same frequency and such that $\dot{\lambda}_i$ does not vanish simultaneously, the carriers follow non-stop elliptical-like paths on the sphere around their cable attachment point. The corresponding velocities of the carriers and  $\dot{\lambda}$ are reported in Figures~\ref{case0_dp} and~\ref{case0_dlambda}, respectively, where it can be appreciated that the norm of the carrier velocities are always larger than a positive constant.

Figure~\ref{fig:case1} and~\ref{fig:case2} show what was referred to as degenerate cases 1 and 2, respectively: 3 carriers maintain the object's pose but one of them stops  along the path. 
In the simulation of case 1, $\forall i,\ \lambda_i(0)=2$ N  $A=1$ N $\psi=1$ rad/s, $\Phi_1=\Phi_3=0$ and $\Phi_2=0.7$ rad. In the simulation of case 2, instead,  $\forall i,\ \lambda_i(0)=2$ N,  $A=1.2$ N, $\psi_1=\psi_3=1$ rad/s, $\psi_2=0.5$ rad/s, $\Phi_1=\Phi_2=0$, and $\Phi_3=0.7$ rad. 
 As it can be seen in Figures~\ref{case1_dlambda} and~\ref{case2_dlambda}, for some time instants  $\dot{\lambda}_1=\dot{\lambda}_j=0$. At the same time instants, the velocity of one carrier (in Figure~\ref{case1_dp} and~\ref{case2_dp}, respectively) goes to zero, too, as predicted by the theory.


%
\section{Conclusions and Future Work}\label{sec:conclusion}
This work studied the possibility of non-stop carrier flights while maintaining a constant pose of cable-suspended objects. After showing that non-stop flights are not feasible for only one or two flying carriers, we showed that three carriers can follow non-stop flights while maintaining the object's pose constant. We also highlighted degenerate cases in which non-stop flight is not achieved even for three carriers. Numerical examples supported the theoretical analysis. 

In the future, we will extend the study to a generic number of carriers $n>0$. We will design an optimal planning algorithm that generates smooth non-stop trajectories for UAVs, explicitly taking into account the system's constraints, such as the minimum forward speed of the robots, their maximum angular velocity, collision avoidance, etc. 
More realistic simulations will be implemented to assess the applicability of the method to VTOL hybrid platforms. Possible mechatronics adaptations to realize experiments will be considered. 
\bibliographystyle{IEEEtran}
\bibliography{biblio}
\end{document}

%% file: symbols.tex

\newcommand\red[1]{{\textcolor{red}{#1}}}
\newcommand\blue[1]{{\textcolor{blue}{#1}}}
\newcommand\green[1]{{\textcolor{green}{#1}}}


\newcommand{\vect}[1]{\bm{#1}}		
\newcommand{\matr}[1]{\bm{#1}}		
\newcommand{\nR}[1]{\mathbb{R}^{#1}}		
\newcommand{\nN}[1]{\mathbb{N}^{#1}}		
\newcommand{\SO}[1]{SO({#1})}		
\newcommand{\define}{:=}			
\newcommand{\modulus}[1]{\left| #1 \right|}	
\newcommand{\matrice}[1]{\begin{bmatrix} #1 \end{bmatrix}}	
\newcommand{\smallmatrice}[1]{\left[\begin{smallmatrix} #1 \end{smallmatrix}\right]}	
\newcommand{\cosp}[1]{\cos \left( #1 \right)}	
\newcommand{\sinp}[1]{\sin \left( #1 \right)}	
\newcommand{\determinant}[1]{\text{det}\left(#1\right)} 	
\newcommand{\sgn}[1]{\text{sgn}\left( #1 \right)}			
\newcommand{\atanTwo}[1]{{\rm atan2}\left( #1\right)}		
\newcommand{\acotTwo}[1]{{\rm acot2}\left( #1\right)}		
\newcommand{\upperRomannumeral}[1]{\uppercase\expandafter{\romannumeral#1}}	
\newcommand{\lowerromannumeral}[1]{\romannumeral#1\relax}
\newcommand{\vSpace}{\;\,}
\newcommand{\image}[1]{\text{Im}\left( #1 \right)}
\newcommand{\pinv}{\dagger}
\newcommand{\diag}[1]{\text{diag}\left( #1 \right)}
\newcommand{\unitOfMeasure}[1]{\; [ \rm #1]}
\newcommand{\Ker}{{\rm Null}}
\newcommand{\ith}[1]{#1^{\rm{th}}}
\newcommand{\fig}{Fig.~}	
\newcommand{\eqn}{Eq.~}	
\newcommand{\tab}{Tab.~}	
\newcommand{\cha}{Chap.~}	
\newcommand{\sect}{Sec.~}	
\newcommand{\theo}{Theorem~}	


\renewcommand{\frame}{\mathcal{F}}		
\newcommand{\origin}{O}						
\newcommand{\vX}{\vect{x}}					
\newcommand{\vY}{\vect{y}}					
\newcommand{\vZ}{\vect{z}}					
\newcommand{\vE}[1]{\vect{e}_{#1}}			
\newcommand{\vV}{\vect{v}}					
\newcommand{\pos}{\vect{p}}				
\newcommand{\dpos}{\dot{\vect{p}}}		
\newcommand{\ddpos}{\ddot{\vect{p}}}	
\newcommand{\rotMat}{\matr{R}}			
\newcommand{\drotMat}{\dot{\matr{R}}}			
\newcommand{\rotMatVectAngle}[2]{\rotMat_{#1}(#2)}	
\newcommand{\angVel}{\vect{\omega}}				
\newcommand{\angAcc}{\dot{\vect{\omega}}}		
\newcommand{\vZero}{\vect{0}}				
\newcommand{\gravity}{\vect{g}}			
\renewcommand{\skew}[1]{\matr{S}(#1)}				
\newcommand{\eye}[1]{\matr{I}_{#1}}		
\newcommand{\roll}{\phi}		
\newcommand{\pitch}{\theta}		
\newcommand{\pitchDes}{\bar{\theta}}		
\newcommand{\pitchEq}{\theta^{eq}}		
\newcommand{\yaw}{\psi}		
\newcommand{\yawDes}{\bar{\psi}}		
\newcommand{\yawEq}{\psi^{eq}}		
\newcommand{\eulerAngles}{\vect{\eta}}

\newcommand{\frameW}{\frame_W}			
\newcommand{\originW}{\origin_W}		
\newcommand{\xW}{\vX_W}				
\newcommand{\yW}{\vY_W}				
\newcommand{\zW}{\vZ_W}				

\newcommand{\frameB}{\frame_B}			
\newcommand{\originB}{\origin_B}			
\newcommand{\xB}{\vX_B}				
\newcommand{\yB}{\vY_B}				
\newcommand{\zB}{\vZ_B}				
\newcommand{\pL}{\pos_L}			
\newcommand{\pLW}{\prescript{W}{}{\pL}} 
\newcommand{\dpL}{\dpos_L}			
\newcommand{\ddpL}{\ddpos_L}		
\newcommand{\rotMatL}{\rotMat_L}
\newcommand{\drotMatL}{\drotMat_L}

\newcommand{\rotMatLEquilib}{\rotMat^{eq}_{L}}
\newcommand{\rotMatLInc}{\hat{\rotMat}_L}
\newcommand{\rotMatLW}{\prescript{W}{}{\rotMat_L}}	
\newcommand{\angVelL}{\angVel_L}		
\newcommand{\angAccL}{\angAcc_L}		
\newcommand{\massL}{{m_L}}
\newcommand{\massLU}{\hat{{m}}_L}				
\newcommand{\inertiaL}{\matr{J}_L}	
\newcommand{\InertiaL}{\matr{M}_L}	
\newcommand{\coriolisL}{\vect{c}_L}	
\newcommand{\gravityL}{\vect{g}_L}	
\newcommand{\graspL}{\matr{G}}		
\newcommand{\dampingL}{\matr{B}_L}	
\newcommand{\f}{\vect{f}}


\newcommand{\length}[1]{{l}_{0#1}}
\newcommand{\lengthU}[1]{\hat{{l}}_{0#1}}	
\newcommand{\springCoeff}[1]{{k}_{#1}}
\newcommand{\springCoeffU}[1]{\hat{{k}}_{#1}}
\newcommand{\cableForce}[1]{\vect{f}_{#1}}
\newcommand{\cableForceEquilib}[1]{\vect{f}^{eq}_{#1}}
\newcommand{\cableForceInc}[1]{\hat{\vect{f}}_{#1}}			
\newcommand{\cableForceU}[1]{\hat{\vect{f}}_{#1}}
\newcommand{\cableAttitudeNorm}[1]{\vect{n}_{#1}}
\newcommand{\cableAttitude}[1]{\vect{l}_{#1}}
\newcommand{\dcableAttitude}[1]{\dot{\vect{l}}_{#1}}	

\newcommand{\condZero}{\xi}
\newcommand{\anchorPoint}[1]{B_{#1}}			
\newcommand{\anchorPos}[1]{\vect{b}_{#1}}		
\newcommand{\anchorLength}[1]{{b}_{#1}}	
\newcommand{\anchorLengthU}[1]{\hat{{b}}_{#1}}		
\newcommand{\anchorPosL}[1]{\prescript{L}{}{\vect{b}}_{#1}}		
\newcommand{\anchorPosLU}[1]{\prescript{L}{}{\hat{\vect{b}}}_{#1}}		

\newcommand{\robotPoint}[1]{A_{#1}}				
\newcommand{\robotPos}[1]{\vect{a}_{ #1}}		
\newcommand{\robotPosP}[1]{\prescript{P}{}{\vect{a}}_{#1}}		
\newcommand{\angleCable}[1]{\alpha_{#1}}		
\newcommand{\angleCables}{\vect{\alpha}}		
\newcommand{\tension}[1]{t_{#1}}				
\newcommand{\tensionMax}[1]{\overline{f}_{L#1}}				
\newcommand{\tensionMin}[1]{\underline{f}_{L#1}}				
\newcommand{\cableForces}{\cableForce{}}		

\newcommand{\frameR}[1]{\frame_{R #1}}			
\newcommand{\originR}[1]{O_{R #1}}					
\newcommand{\xR}[1]{\vX_{R #1}}								
\newcommand{\yR}[1]{\vY_{R #1}}								
\newcommand{\zR}[1]{\vZ_{R #1}}								
\newcommand{\pR}[1]{\pos_{R #1}}						
\newcommand{\dpR}[1]{\dpos_{R #1}}					
\newcommand{\ddpR}[1]{\ddpos_{R #1}}				
\newcommand{\uR}[1]{\vect{u}_{R #1}}				
\newcommand{\pRW}[1]{\prescript{W}{}{\pR{#1}}} 	
\newcommand{\rotMatR}[1]{\rotMat_{R #1}}			
\newcommand{\thrust}[1]{\vect{f}_{R #1}}		
\newcommand{\maxThrust}[1]{h_{#1}}				
\newcommand{\thrustIJ}{\thrust{ij}}				
\newcommand{\maxThrustIJ}{\maxThrust{ij}}		
\newcommand{\gravityIJ}{\vect{g}_{ij}}			
\newcommand{\massR}[1]{m_{R#1}}					
\newcommand{\inertiaR}[1]{\vect{J}_{R#1}}		

\newcommand{\dampingA}[1]{\matr{B}_{A#1}}		
\newcommand{\springA}[1]{\matr{K}_{A#1}}		
\newcommand{\inertiaA}[1]{\matr{M}_{A#1}}		
\newcommand{\uA}[1]{\vect{u}_{A#1}}					
\newcommand{\paramA}[1]{\vect{\pi}_{A#1}}			

\newcommand{\config}{\vect{q}}					
\newcommand{\dconfig}{\vect{v}}			
\newcommand{\ddconfig}{\dot{\dconfig}}			
\newcommand{\configR}{\config_R}					
\newcommand{\dconfigR}{\dconfig_R}				
\newcommand{\ddconfigR}{\ddconfig_R}			
\newcommand{\configL}{\config_L}					
\newcommand{\dconfigL}{\dconfig_L}				
\newcommand{\ddconfigL}{\ddconfig_L}			
\newcommand{\state}{\vect{x}}						
\newcommand{\dynamicModelFun}{m}					

\newcommand{\configEq}{\bar{\config}}			
\newcommand{\configLEq}{\bar{\config}_L}		
\newcommand{\configREq}{\bar{\config}_R}		
\newcommand{\paramAEq}[1]{\bar{\vect{\pi}}_{A#1}}			
\newcommand{\paramAEqInc}[1]{\hat{\bar{\vect{\pi}}}_{A#1}}
\newcommand{\paramAEqU}[1]{\hat{\bar{\vect{\pi}}}_{A#1}}
\newcommand{\pLEq}{\bar{\pos}_L}

\newcommand{\pLEquilib}{{\pos}^{eq}_L}
\newcommand{\rotMatLEq}{\bar{\rotMat}_L}		
\newcommand{\paramASetEq}{{\Pi}_{A}(\configLEq)}				
\newcommand{\paramASetEqPrime}{{\Pi}_{A}(\configLEq')}				
\newcommand{\configRSetEq}{\mathcal{P}_{R}}			
\newcommand{\cableForceEq}[1]{\bar{\vect{f}}_{#1}}
\newcommand{\cableForceEqInc}[1]{\hat{\bar{\vect{f}}}_{#1}}				
\newcommand{\cableForcesEq}{\bar{\vect{f}}}	
\newcommand{\cableForcesEqInc}{\hat{\bar{\vect{f}}}}					
\newcommand{\cableForcesSetEq}{\mathcal{F}_{L}}		
\newcommand{\nullGrasp}{\vect{r}_L}				
\newcommand{\internalTension}{t_L}				
\newcommand{\internalForceDir}{\vect{n}_L}	
\newcommand{\internalForceDirL}{\prescript{L}{}{\vect{n}}_L}	
\newcommand{\pREq}[1]{\bar{\pos}_{R #1}}	
\newcommand{\pREqInc}[1]{\hat{\bar{\pos}}_{R #1}}		

\newcommand{\pREquilib}[1]{\pos^{eq}_{R #1}}
\newcommand{\configRparamASetEq}{\mathcal{S}(\configLEq)}
\newcommand{\configRparamAEq}{\bar{\vect{s}}}
\newcommand{\configRparamA}{{\vect{s}}}
\newcommand{\configSetEq}{\mathcal{Q}(\internalTension,\configLEq)}
\newcommand{\configSetEqZero}{\mathcal{Q}(0,\configLEq)}
\newcommand{\configSetEqZeroi}[1]{\mathcal{Q}_{#1}(0,\configLEq)}

\newcommand{\configSetEqPlus}{\mathcal{Q}^+(\internalTension,\configLEq)}
\newcommand{\configSetEqMinus}{\mathcal{Q}^-(\internalTension,\configLEq)}

\newcommand{\configRLSetEq}{\mathcal{R}(\internalTension,\configLEq)}
\newcommand{\configRLSetEqZero}{\mathcal{R}(0,\configLEq)}

\newcommand{\errorpREq}[1]{\vect{e}_{R#1}}	
\newcommand{\errorPL}{{\vect{e}_{p}}_L}

\newcommand{\screwJacobian}{\matr{J}(\genCoord)}		
\newcommand{\vectI}{\vect{v}_i}
\newcommand{\vectII}{{v}_{i,i}}
\newcommand{\minTensionResolving}{\underline{t}_i(\genCoord)}		
\newcommand{\maxTensionResolving}{\overline{t}_i(\genCoord)}		

\newcommand{\stateEq}{\bar{\state}}
\newcommand{\invariantSet}{\Omega_{\alpha}}		
\newcommand{\invariantSetZero}{\Omega_{0}}		
\newcommand{\dVZeroSet}{\mathcal{E}}				
\newcommand{\stateSetEq}{\mathcal{X}(\internalTension,\configLEq)}
\newcommand{\stateSetEqZero}{\mathcal{X}(0,\configLEq)}
\newcommand{\stateSetEqZeroi}[1]{\mathcal{X}_{#1}(0,\configLEq)}
\newcommand{\stateSetEqPlus}{\mathcal{X}^+(\internalTension,\configLEq)}
\newcommand{\stateSetEqMinus}{\mathcal{X}^-(\internalTension,\configLEq)}
\newcommand{\stateSetEqPlusPrime}{\mathcal{X}^+(\internalTension',\configLEq)}
\newcommand{\lyapunovFun}{V(\state)}				

\newcommand{\Vadd}{V_R(\state)}				
\newcommand{\dVadd}{\dot{V}_1(\state)}	

\newcommand{\dlyapunovFun}{\dot{V}(\state)}				
\newcommand{\maxInvariantSet}{\mathcal{M}}

\newcommand{\inp}{\vect{u}}
\newcommand{\out}{\vect{y}}
\newcommand{\outputFunction}{\vect{\Phi}(\out)}

\newcommand{\displacement}{\vect{d}}
\newcommand{\pREqIncRef}[1]{\pREqInc{#1}^r}
\newcommand{\rotMatRRef}[1]{\rotMatR{#1}^r}

\newtheorem{problem}{Problem}
\newtheorem{prop}{Proposition}
\newtheorem{lemma}{Lemma}

\newtheorem{fact}{Fact}

\theoremstyle{remark}
\newtheorem{remark}{Remark}